\documentclass{article}

\usepackage{microtype}
\usepackage{graphicx,amsmath,amssymb,cite}
\usepackage{subfigure}
\usepackage{booktabs} 

\usepackage{hyperref}

\usepackage[accepted]{icml2020}

\newcommand\numberthis{\addtocounter{equation}{1}\tag{\theequation}}

\usepackage{amsfonts}
\graphicspath{ {./figures/} }
\usepackage{comment}
\usepackage{lipsum}
\usepackage{mwe}
\usepackage{caption}
\usepackage{graphicx}
\usepackage{setspace}

\icmltitlerunning{Quantized Proximal Averaging Network for Analysis Sparse Coding}
\onecolumn

\begin{document}


\twocolumn[
\icmltitle{Quantized Proximal Averaging Network for Analysis Sparse Coding}




\icmlsetsymbol{equal}{*}

\begin{icmlauthorlist}
\icmlauthor{Kartheek Kumar Reddy Nareddy*}{ee}
\icmlauthor{Mani Madhoolika Bulusu*}{ee}
\icmlauthor{Praveen Kumar Pokala}{ee}
\icmlauthor{Chandra Sekhar Seelamantula}{ee}
\end{icmlauthorlist}

\icmlaffiliation{ee}{Electrical Engineering, Indian Institute of Science}

\icmlcorrespondingauthor{Kartheek Kumar Reddy Nareddy}{\mbox{nareddyreddy@iisc.ac.in}}

\vskip 0.3in
]

\printAffiliationsAndNotice{\icmlEqualContribution}




\begin{abstract}
	
We solve the analysis sparse coding problem considering a combination of convex and non-convex sparsity promoting penalties. The multi-penalty formulation results in an iterative algorithm involving {\it proximal-averaging}. We then unfold the iterative algorithm into a trainable network that facilitates learning the sparsity prior. We also consider quantization of the network weights. Quantization makes neural networks efficient both in terms of memory and computation during inference, and also renders them compatible for low-precision hardware deployment. Our learning algorithm is based on a variant of the ADAM optimizer in which the quantizer is part of the forward pass and the gradients of the loss function are evaluated corresponding to the quantized weights while doing a book-keeping of the high-precision weights. We demonstrate applications to compressed image recovery and magnetic resonance image reconstruction. The proposed approach offers superior reconstruction accuracy and quality than state-of-the-art unfolding techniques and the performance degradation is minimal even when the weights are subjected to extreme quantization.
	

\end{abstract}

\section{Introduction}
The objective in Compressive Sensing (CS) is to reconstruct a sparse signal ${\boldsymbol x}\in \mathbb{R}^n$ from its compressed observation, $\boldsymbol{y = \Phi x} \in \mathbb{R}^m$, where $\mathbf{\Phi}$ is an ${m \times n}$ matrix, with $ m \ll n$. The standard convex optimization formulation to solve the CS problem is given as follows:
\begin{equation}
\underset{\boldsymbol x}{\min} ~~\frac{1}{2}\|\boldsymbol{\Phi x - y}\|^2_2 + \lambda \|\boldsymbol{\Psi x}\|_1, \label{eq: CS equation}
\end{equation}
where $\boldsymbol{y}$ is the observation, $\boldsymbol{\Psi}$ is the sparsifying transform for $\boldsymbol{x}$, and sparsity is enforced on $\boldsymbol{\Psi x}$ using an appropriate regularizer, which in this case is the $\ell_1$-norm, and $\lambda$ is the regularization parameter.

However it has been shown that $\ell_1$-norm regularization results in biased amplitude estimates \cite{mcp}. Non-convex regularizers are known to reduce the bias in the estimates. 
One can design efficient non-convex regularizers by considering a combination of several sparsity-promoting regularizers.

\begin{figure}[t]
		\centering
		\includegraphics[width=1.0\linewidth]{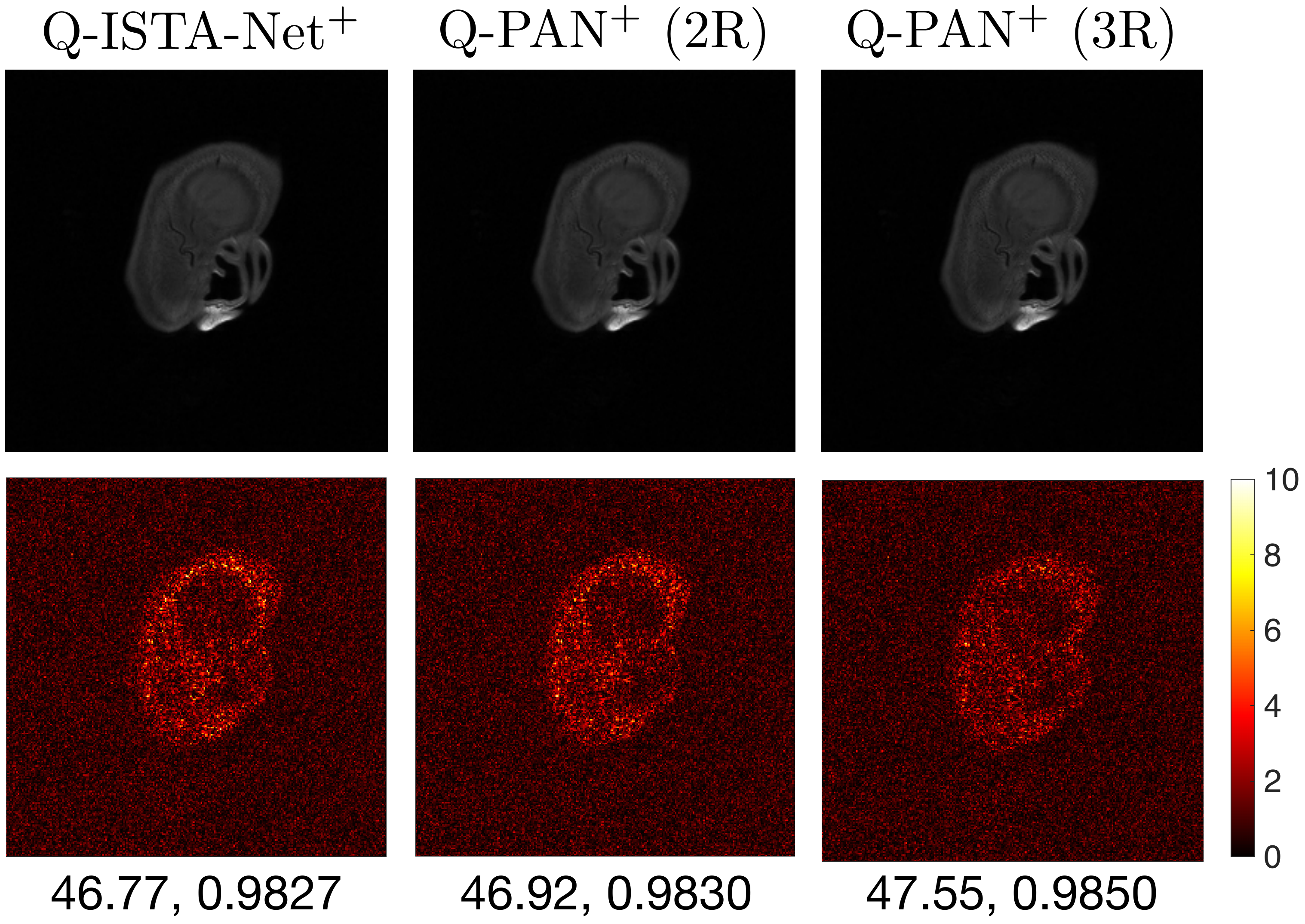}
	\caption{Brain MRI image reconstruction for CS ratio 40\% by the proposed 1-bit Quantized models. The bottom row visualizes the absolute difference between the ground truth and the reconstructed images. The Q-PAN$^+$ (3R) model captured the brain structure better and hence the difference image is majorly composed of noise.The numbers indicates the PSNR, SSIM values.}
	\label{img:brain-mri-reconstructions}
\end{figure}
The CS framework has enabled several practical applications particularly in the context of computational imaging \cite{CS-image}, ultrasound imaging \cite{CS-ultra-sound}, magnetic resonance imaging \cite{CS-MRI}, cognitive radar \cite{CS-Cog-Radar}, cognitive radio \cite{CS-Cog-Radio}, etc.\\
\indent Sparse recovery algorithms are iterative and involve manual parameter tuning and several iterations to achieve accurate reconstruction. Iterative algorithms when unfolded into a feedforward neural network architecture can greatly enhance the reconstruction accuracy since the network has the advantage of {\it learnability}. Further, the depth of the network is also significantly lower than the number of iterations of the algorithm from which it is inspired. Therefore, the inference time is reduced. However, the price to pay for these advantages is the overhead of training.
\indent A practical bottleneck in deploying such models is that they require more memory to store the high-precision weights and consequently also more computation. In applications involving low-precision hardware, the reconstruction accuracy reduces drastically. One of the issues that is addressed in this paper is the effect of weight quantization on the performance of unfolded networks. The objective is to obtain comparable accuracy to the full-precision scenario considering weights that are coarsely quantized (which we show in the Figure~\ref{img:brain-mri-reconstructions}). Such {\it quantized unfolded networks} are more amenable to practical implementation particularly in resource constrained settings.\\
\indent Another important aspect that we focus on in this paper is the problem of {\it analysis prior learning} using a combination of convex/non-convex sparsity promoting  regularizers. The analysis prior is more effective than the synthesis prior because most naturally occurring signals and images are sparse only when analyzed using a suitable transformation \cite{Elad-Analysis}. Further, employing a combination of penalty functions has been shown to result in superior reconstruction performance than a single penalty function \cite{Kamilov-proximal} and gives rise to the notion of {\it proximal averaging}. The penalty functions could be convex or non-convex. {\it Learning} a convex combination of penalty functions is also likely to improve the sparse recovery performance.\\
\indent Before proceeding with further developments, we review recent state of the art in these domains. 

\begin{table*}[t]
\caption{Sparsity-promoting regularizers and corresponding proximal operators employed in this study.}
\label{table:regularizers}
\vskip 0.15in
\begin{center}
\begin{small}
\begin{sc}
\resizebox{\linewidth}{!}{
\begin{tabular}{p{0.189\linewidth}|c|c}
\toprule
Name & Penalty Function & Proximal Operator\\
\midrule
$\ell_1-\mathrm{norm}$ $(\lambda>0)$ & $g_1(x)=\lambda \left| x \right|$ & $\mathcal{P}_{1}(x) = \mathrm{sgn}(x) \mathrm{max}\left(\left| x \right| - \lambda, 0\right)$\\
\midrule
MCP \cite{mcp} $(\lambda>0, \gamma>1)$ &%
$g_2(x)=\begin{cases} {\lambda  \left| x \right| - \cfrac{{\left| x \right|}^2}{2 \lambda  \gamma }, } \qquad\enspace\, {\mathrm{for} \left| x \right| \leq \gamma\lambda ,} \\
{\cfrac{{\lambda }^{2}\gamma }{2}, } \qquad\qquad\qquad {\mathrm{for} \left| x \right| \geq \gamma  \lambda .} \end{cases}$ &%
$\mathcal{P}_{2}(x)=\begin{cases} {0,} \qquad\qquad\qquad\qquad\qquad {\mathrm{for} \; \left| x \right| \leq \lambda,} \\
{\mathrm{sgn} \left(x\right) \cfrac{\gamma}{\gamma - 1} \left( \left| x \right| - \lambda \right),} \enspace\, {\mathrm{for} \; \lambda < \left| x \right| \leq \gamma \lambda,} \\
{x,} \qquad\qquad\qquad\qquad\qquad {\mathrm{for} \; \left| x \right| > \gamma \lambda .} \end{cases}$ \\%
\midrule
SCAD \cite{scad} $(\lambda>0, a>2)$ &%
$g_3(x)=\begin{cases} {\lambda  \left| x \right|,} \qquad\qquad\qquad\quad {\mathrm{for} \left| x \right| \leq \lambda ,} \\
{\cfrac{{\left| x \right|}^{2} - 2a\lambda \left| x \right| + \lambda ^{2}}{2(1-a)},} \enspace\, {\mathrm{for} \lambda  < \left| x \right| \leq a\lambda ,} \\
{\cfrac{(a+1){\lambda ^{2}}}{2},} \qquad\qquad\enspace\, {\mathrm{for} \left| x \right| > a\lambda .} \end{cases}$ &%
$\mathcal{P}_{3}(x)=\begin{cases} {\mathrm{sgn}\left(x\right) \mathrm{max} \left( \left| x \right| - \lambda, 0 \right),} \, {\mathrm{for} \; \left| x \right| \leq 2\lambda,} \\
{\cfrac{\left(a - 1\right)x - \mathrm{sgn} \left(x\right) a \lambda}{a-2},} \enspace\, {\mathrm{for} \; 2\lambda < \left| x \right| \leq a \lambda,} \\
{x,} \qquad\qquad\qquad\qquad\qquad {\mathrm{for} \; \left| x \right| > a \lambda .} \end{cases}$ \\
\bottomrule
\end{tabular}}
\end{sc}
\end{small}
\end{center}
\end{table*}

\section{Prior Art}

We present the literature in two categories, namely, quantized networks and analysis sparse recovery techniques.

\textit{Quantized Networks}
Neural Networks that employ binary weights and activations, also known as binary neural networks (BNNs) \cite{Binaryconnect,BNN,JMLR-QNN} have been shown to be promising for solving classification problems. Their efficacy has been demonstrated on MNIST \cite{MNIST} and CIFAR10 \cite{CIFAR10} datasets. The XNOR-Net \cite{XNOR-Net} used a gain term computed from the statistics of the weights to scale the binary weights in order to reduce the quantization error. The XNOR-Net performed relatively better than previous BNN architectures on the ImageNet \cite{imagenet} dataset. The learned quantization networks (LQ-Net) \cite{LQ-Net} was proposed to minimize the quantization error by jointly learning the quantizer and the network. The Bi-Real Network \cite{bi-real} employs residual connections where full-precision activations from the previous layer affect the quantization in the current layer. On the optimization front \citet{bayesian-BNN} justified the straight-through-estimator (STE) method used for training the BNNs through bayesian learning. The quantization of weights has also been shown to be promising in the context of generative adversarial networks (GANs) \cite{Q-GAN,deep-Q-GAN}. \citet{NASB} adapts Neural Architecture Search (NAS) to find an optimized architecture for the binarization of Convolutional Neural Networks (CNNs). \citet{robust-QNN} showed that quantized neural networks are generally robust relative to their full precision counterpart against adversarial perturbations.
While the role of quantization has been considered in classification and generation problems, it has been relatively unexplored in the context of optimizing unfolded networks for sparse recovery. 

\textit{Analysis Sparse Priors} Several works \cite{beck2009fast,elad2007analysis,liu2016projected, AMP} have considered the $\ell_1$-regularized analysis/synthesis sparse recovery problems owing to its convexity. In order to solve the CS problem in the context of recovering natural images, optimization based techniques assumed that the input images are sparse in a fixed transform domain such as the discrete cosine transform (DCT), wavelet domain \cite{wavelet-domain}, gradient-domain \cite{TVAL3}, etc. Denoising based AMP (D-AMP) \cite{D-AMP}, etc. ensures good recovery for natural images. Most methods exploit structured sparsity of the data as an image prior and solve the problem iteratively \cite{CS-with-GMM, TVAL3, D-AMP, adaptive-sparsifying-basis}. All iterative sparse recovery methods require hundreds of iterations to solve Eq.~\eqref{eq: CS equation}, which inevitably gives rise to high computational overhead thus restricting the application of CS. In addition, the analysis operator or the hyperparameters such as learning rate and regularization parameter are usually hand-crafted.

To overcome the limitations of optimization based techniques, network-based approaches have been proposed recently. ReconNet is a neural network approach proposed in \cite{ReconNet}, which takes in CS measurements of an image as input and outputs an intermediate reconstruction. Networks such as ISTA-Net \cite{ISTA-Net} performed better when the sparsifying transform was learned by imposing the $\ell_1$ penalty on the prior. 
Two aspects are missing from current literature. First, finer approximation of $\ell_0$ pseudonorm in the context of data-driven analysis-prior learning and second, reducing the model size and improved hardware-compatibility of the CS networks. We aim to address these two aspects with application to compressed image reconstruction.


\section{Contribution of this Paper}
\indent Our contribution lies at the intersection between the two aspects -- network quantization and optimization for analysis sparse prior learning while employing several regularizers.\\
\indent The starting point for the developments is a multi-penalty/composite regularization formulation of the sparse recovery problem in an analysis-sparse setting. The result is the proximal-averaged iterative shrinkage algorithm (PAISA), which when unfolded gives rise to the proximal-averaging network (PAN). It is in the context of PAN that we consider the effect of quantization (Q-PAN). On the application front, we consider image recovery from compressed measurements both for natural images and magnetic resonance images. 
\begin{figure*}[t]
\vskip 0.2in
\begin{center}
    \centerline{\includegraphics[width=0.9\linewidth]{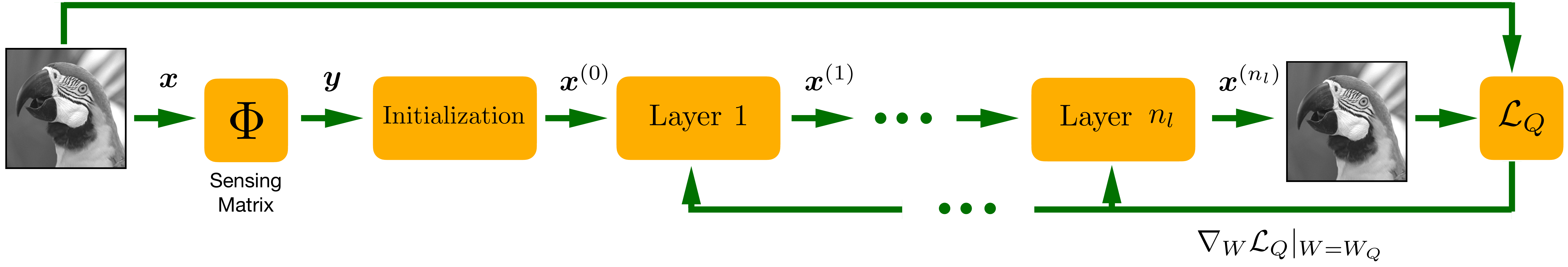}}
    \caption{Illustration of the proposed Quantized Proximal Averaging Network (Q-PAN).}
    \label{pipeline}
\end{center}
\vskip -0.2in
\end{figure*}

\section{Composite Regularization in the Analysis Setting}
\label{sec-PAN}
Consider the problem of compressed sensing recovery based on multiple penalties $\{g_i\}$ as given below:
\begin{align*}
    \underset{\boldsymbol x}{\min} & ~\frac{1}{2}\|\mathbf{\Phi}\boldsymbol x - \boldsymbol y\|^2_2 + \underbrace{\sum_{i=1}^{p} \alpha_i g_i(\mathcal{F}(\boldsymbol{x}))}_{g(\boldsymbol{x})} \\
    & \text{subject to } \sum_{i=1}^{p} \alpha_i = 1, \quad 0 < \alpha_i < 1, \forall i, \numberthis 
    \label{eq: cs-composite}
\end{align*}
where $\mathcal{F}$ is the data-driven, sparsifying, analysis operator and $p$ is the number of sparsity-promoting regularizers. The convex and non-convex regularizers that are considered in our work are the $\ell_1$-norm, minimax-concave penalty (MCP) \cite{mcp}, and smoothly clipped absolute deviation (SCAD) \cite{scad}. While the $\ell_1$-norm is convex, the other two penalties -- MCP and SCAD -- are not convex. The MCP and SCAD penalties approximate the $\ell_0$ pseudonorm better than the $\ell_1$ norm and hence give rise to better recovery performance \cite{mcp}. The ideal sparse recovery problem requires minimization of the $\ell_0$ pseudonorm, but it is NP-hard and hence tractable solvers that approximate $\ell_0$ minimization are preferred. The MCP and SCAD penalties perfectly fit the bill and hence we consider them for composite regularization. The three penalties under consideration and the corresponding proximal operators are given in Table \ref{table:regularizers}.

\subsection{Proximal-Averaged Iterative Shrinkage Algorithm (PAISA)}
\label{sec:paisa}
The problem in Eq.~\eqref{eq: cs-composite} is optimized based on majorization-minimization \cite{mm-technique} and proximal-averaging \cite{proximal-avg}. The update $\boldsymbol{x}$ at $(k+1)^{\text{th}}$ iteration is obtained by solving the following optimization problem:
\begin{equation}
    \underset{\boldsymbol{x}}{\min} ~\frac{1}{2\rho}\|\boldsymbol{x} - \boldsymbol{r}^{(k+1)}\|^2_2 + \sum_{i=1}^{p} \alpha_i g_i(\mathcal{F}(\boldsymbol{x})),
\end{equation}
where $\boldsymbol{r}^{(k+1)} = \boldsymbol{x}^{(k)} - \rho {\mathbf{\Phi}^\textsc{T} (\mathbf{\Phi} \boldsymbol{x}^{(k)} - \boldsymbol{y})}$ and $\rho$ is the step-size. The above problem can be rewritten as
\begin{equation}
    \underset{\boldsymbol x}{\min} ~ {\sum_{i=1}^{p} \left(\frac{\alpha_i}{2\rho}\|\boldsymbol{x} - \boldsymbol{r}^{(k+1)}\|^2_2 +  \alpha_i g_i(\mathcal{F}(\boldsymbol{x}))\right)}.  \label{update-eqn-composite}
\end{equation}
 \citet{ISTA-Net} showed that a data-driven analysis transform improves the reconstruction  of natural images from the compressed measurements. Based on this observation, $\mathcal{F}$ is chosen to be a combination of two linear convolutional operators, without bias terms, separated by a rectified linear unit (ReLU) \cite{relu}. $\mathcal{F}$ can be formulated in matrix form as  $\mathcal{F}(\boldsymbol{x}) = \mathbf{B}\,\text{max}(\mathbf{A}\boldsymbol{x}, \boldsymbol{0})$, where $\mathbf{A}$ and $\mathbf{B}$ correspond to the two convolutional operators, and $\text{max}(\cdot,\boldsymbol{0})$ denotes the ReLU. The convolutional operators $\mathbf{A}$ and $\mathbf{B}$ use $n_f$ filters of size $3 \times 3$ and $3 \times 3 \times n_f$, respectively.
 \citet{group-based-sparse-rep} showed that, in the context of inverse problems in imaging, it is reasonable to assume that the elements of $(\boldsymbol{x}^{(k+1)} - \boldsymbol{r}^{(k+1)})$ are independent normal distributed random variables with zero mean and variance $\sigma^2$. Let $\boldsymbol{r}^{(k+1)}$ and  $\mathcal{F}(\boldsymbol{r}^{(k+1)})$ denote the mean values of $\boldsymbol{x}$ and $\mathcal{F}(\boldsymbol{x})$, respectively. The following approximation holds \cite{ISTA-Net}:
\begin{equation}
\label{approximation}
\|\mathcal{F}(\boldsymbol{x})-\mathcal{F}(\boldsymbol{r}^{(k+1)})\|^2_2 \approx \beta \|\boldsymbol{x}-\boldsymbol{r}^{(k+1)}\|^2_2,
\end{equation}
where $\beta$ depends only on the parameters of $\mathcal{F}$. Incorporating the approximation in Eq.~\eqref{approximation} into Eq.~\eqref{update-eqn-composite}, we obtain (merging $\rho$ and $\beta$ into the parameters of the regularizers):
\begin{eqnarray}
\boldsymbol{x}^{(k+1)} &=& \text{arg} \min_{\boldsymbol{x}} \sum_{i=1}^{p} \alpha_i\|\mathcal{F}(\boldsymbol{x})-\mathcal{F}(\boldsymbol{r}^{(k+1)})\|^2_2  \nonumber \\ &&+ \sum_{i=1}^{p} \alpha_i g_i(\mathcal{F}(\boldsymbol{x})) \label{eq: paisa-opt-prob}.
\end{eqnarray}
The solution to the above problem relies on proximal-averaging and is given by
\begin{equation}
\label{eq: paisa-update-step}
    \boldsymbol{x}^{(k+1)}
    = \widetilde{\mathcal{F}}
    \Bigg(
    \sum_{i=1}^{p} \alpha_i \mathcal{P}_i(\mathcal{F}(\boldsymbol{r}^{(k+1)}))
    \Bigg),
\end{equation}
where $\widetilde{\mathcal{F}}$ is the adjoint of $\mathcal{F}$ such that
$\widetilde{\mathcal{F}}( \mathcal{F}(\boldsymbol{x})) = \boldsymbol{x}$. The optimization steps for minimizing the problem in Eq.~\eqref{eq: cs-composite} are listed in Algorithm~\ref{alg:paisa}. 

\subsection{Proximal Averaging Network (PAN)}
Unfolding the iterations of PAISA (Algorithm~\ref{alg:paisa}) results in a feedforward neural network, which we refer to as the proximal averaging network (PAN). A single PAN layer is represented by
\begin{equation}
\label{eq: pan-layer}
    \boldsymbol{x}^{(k)}
    = \widetilde{\mathcal{F}}^{(k)}
    \Bigg(
    \sum_{i=1}^{p} \alpha_i \mathcal{P}_i(\mathcal{F}^{(k)}(\boldsymbol{r}^{(k)}))
    \Bigg).
\end{equation}
The parameters corresponding to the sparsity promoting regularizers and the analysis transform $\{\lambda_1^{(k)}, \lambda_2^{(k)}, \lambda_3^{(k)}, a^{(k)}, \gamma^{(k)}, \mathcal{F}^{(k)},  \widetilde{\mathcal{F}}^{(k)}\}$ are all learnable subject to the conditions $\lambda_i^{(k)} > 0, \gamma^{(k)} > 1, a^{(k)} > 2$ for each layer $k$ of the network.

\begin{algorithm}[tb]
   \caption{Proximal-Averaged Iterative Shrinkage Algorithm (PAISA)}
   \label{alg:paisa}
\begin{algorithmic}
   \STATE {\bfseries Input:} $\boldsymbol{x}$, K, $\boldsymbol{\lambda}_1$, $\boldsymbol{\lambda}_2$, $\boldsymbol{\lambda}_3$, $\boldsymbol{\gamma}$, $\boldsymbol{a}$
   \STATE Initialize $\boldsymbol{x}_0, k = 0$.
   \WHILE{$k \leq K-1$}
       \STATE $\boldsymbol{x}_k = \widetilde{\mathcal{F}}
              (\sum_{i=1}^{p} \alpha_i \mathcal{P}_i(\mathcal{F}(\boldsymbol{r}_{k}))) $
       \STATE $k \gets k+1$
   \ENDWHILE
\end{algorithmic}
\end{algorithm}

\begin{figure*}[t]
\vskip 0.2in
\begin{center}
    \centerline{\includegraphics[width=0.95\linewidth]{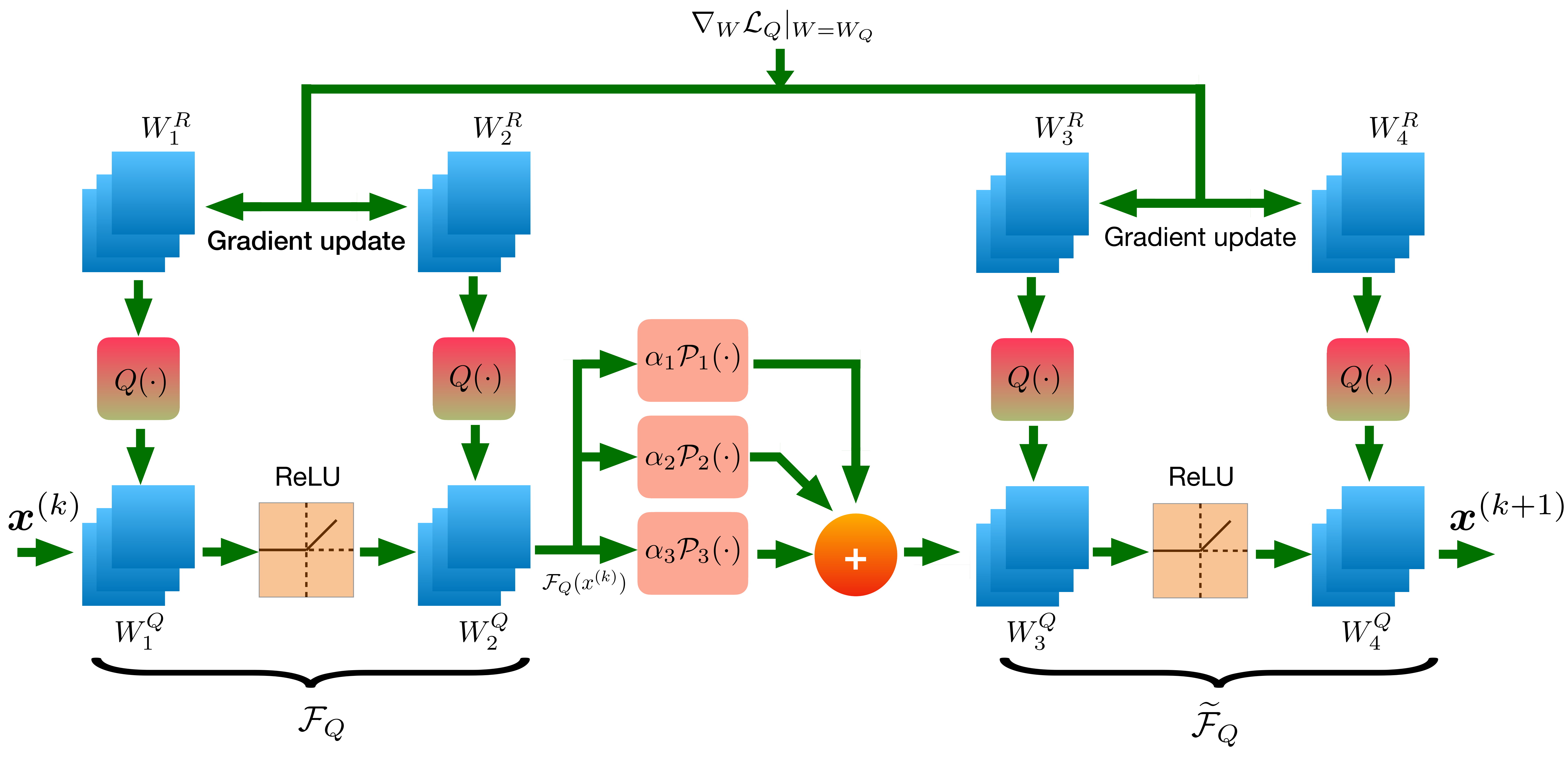}}
    \caption{Structure of a layer in the Q-PAN. Quantized weights ($\mathbf{W}_i^Q$) are obtained from their full precision copy ($\mathbf{W}_i^R$) via quantization operator $Q(\cdot)$. The intermediate reconstruction $\boldsymbol{x}^{(k)}$ is processed through convolution filters $\{\mathbf{W}_i^Q\}$, ReLU transform, proximal operators $\{\mathcal{P}_i\}$ to give $\boldsymbol{x}^{(k+1)}.$}
    \label{recon-block}
\end{center}
\vskip -0.2in
\end{figure*}

\subsection{Loss Function}
The loss function comprises two terms. The first term captures the mean-squared error between the ground truth $\boldsymbol{x}_i$ and the reconstructed image $\boldsymbol{x}^{(n_l)}_i$, where $n_l$ denotes the number of layers in the network. The second term seeks to enforce invertibility of the sparsifying transform $\mathcal{F}$ across the layers. The loss function $\mathcal{L}$ proposed in \cite{ISTA-Net} is used in our training:
\begin{eqnarray}
\mathcal{L} &=&\frac{1}{N}{\sum^{n_b}_{i=1}\|\boldsymbol{x}^{(n_l)}_i - \boldsymbol{x}_i\|^2_2}  \\
&&+\gamma\, \frac{1}{N}\, {\sum^{n_b}_{i=1}\sum^{n_l}_{k=1}\|\widetilde{\mathcal{F}}^{(k)}(\mathcal{F}^{(k)}(\boldsymbol{x}_i)) - \boldsymbol{x}_i\|^2_2}, \label{loss-function} \nonumber
\end{eqnarray}
where $n_b$ is the number of training patches extracted from the images in the dataset, $n$ is the size of each training patch, $n_l$ is the number of layers of the network, $N = n_bn$, and $\gamma$ determines the trade-off between the two terms under consideration. $\gamma$ is fixed at 0.01 in our experiments. 

\subsection{PAISA$^+$ Update}
The residuals (defined as the difference between the ground truth and its prediction) of natural images and videos are compressible \cite{jpeg, video-coding}. The residual learning framework named ResNet \cite{resnet} explicitly formulates the layers of the network to learn residual functions with reference to the layer inputs. Drawing inspiration from ResNet and ISTA-Net$^+$, we propose PAISA$^+$, whose iterations are unfolded to yield PAN$^+$, which learns a residual function for the update $\boldsymbol{x}_{k+1}$, instead of learning it directly as done in PAN.

\indent The update step is: $\boldsymbol{x}^{(k+1)} = \boldsymbol{r}^{(k+1)} + \boldsymbol{w}^{(k+1)} + \boldsymbol{e}^{(k+1)}$, where $\boldsymbol{w}^{(k+1)}$ is the residual and $\boldsymbol{e}^{(k+1)}$ is the error, as recommended in ISTA-Net$^+$ \cite{ISTA-Net}. The residual $\boldsymbol{w}^{(k+1)}$ contains the high-frequency component of $\boldsymbol{x}^{(k+1)}$ missing from $\boldsymbol{r}^{(k+1)}$. It can be extracted from $\boldsymbol{x}^{(k+1)}$ by an affine transformation $\mathcal{R}$ i.e., $\boldsymbol{w}^{(k+1)}$ = $\mathcal{R}(\boldsymbol{x}^{(k+1)})$ = $\mathcal{G}(\mathcal{D}(\boldsymbol{x}^{(k+1)}))$, where $\mathcal{D}$ has $n_f$ filters (of size $3 \times 3$) and $\mathcal{G}$ has one filter (of size $3 \times 3 \times n_f$).

To obtain a closed-form solution, we model $\mathcal{F} = \mathcal{H} \circ \mathcal{D}$, where $\mathcal{H}$ is composed of two convolutional operators separated by a ReLU. Substituting $\mathcal{F}$ in Eq.~\eqref{eq: paisa-opt-prob} by $\mathcal{H} \circ \mathcal{D}$ gives
\begin{eqnarray}
    \boldsymbol{x}^{(k+1)} &=& \text{arg} \min_{\boldsymbol{x}} ~~\frac{1}{2}\|\mathcal{H}(\mathcal{D}(\boldsymbol{x}))-\mathcal{H}(\mathcal{D}(\boldsymbol{r}^{(k+1)}))\|^2_2 \nonumber \\ &&+ \sum_{i=1}^{p} \alpha_i g_i(\mathcal{H}(\mathcal{D}(\boldsymbol{x}))). \nonumber
\end{eqnarray}
Following the same strategy as in PAISA, we define $\widetilde{\mathcal{H}}$, the adjoint of $\mathcal{H}$ to be symmetric. The PAISA$^+$ update is:
\begin{eqnarray}
\boldsymbol{x}^{(k+1)} &=& \boldsymbol{r}^{(k+1)} \label{eq: paisa-plus-update}\\
&&+\, \mathcal{G}\bigg(\widetilde{\mathcal{H}}\bigg(\sum_{i=1}^{p} \alpha_i \mathcal{P}_i(\mathcal{H}(\mathcal{D}(\boldsymbol{r}^{(k+1)})))\bigg)\bigg). \nonumber
\end{eqnarray}
The iterations of PAISA$^+$ are unfolded into the layers of a neural network, henceforth referred to as PAN$^+$. The parameters in all the layers of the model are learnable, i.e., all convolutional operators $\mathcal{H}$, $\widetilde{\mathcal{H}}$, $\mathcal{D}$, $\mathcal{G}$ are learned in each layer of the network, along with parameters involved in the proximal operators $\mathcal{P}_i$. A layer of PAN$^+$ is given by the input-output equation:
\begin{eqnarray}
\boldsymbol{x}^{(k)} &=& \boldsymbol{r}^{(k)} \label{eq: paisa-plus-layer}\\
&& \hspace{-0.5cm} +\,  \mathcal{G}^{(k)}\bigg(\widetilde{\mathcal{H}}^{(k)}\bigg(\sum_{i=1}^{p} \alpha_i \mathcal{P}_i(\mathcal{H}^{(k)}(\mathcal{D}^{(k)}(\boldsymbol{r}^{(k)})))\bigg)\bigg). \nonumber
\end{eqnarray}

The learnable parameters in this case are \{$\lambda_1^{(k)}, \lambda_2^{(k)}, \lambda_3^{(k)},\\ a^{(k)}, \gamma^{(k)}, \mathcal{H}^{(k)}, \widetilde{\mathcal{H}}^{(k)}, \mathcal{D}^{(k)}, \mathcal{G}^{(k)}$\} subject to the conditions $\lambda_i^{(k)} > 0, \gamma^{(k)} > 1, a^{(k)} > 2$  for each layer $k$ of the network. Also, the loss function is augmented with a new term as done in Eq.~\eqref{loss-function} to enforce the adjoint property: $\widetilde{\mathcal{H}}^{(k)}\circ \mathcal{H}^{(k)}=\mathcal{I}$.

\begin{algorithm}[htb]
	\caption{Training strategy for Q-PAN}
	\label{alg:Q-PAN-Train}
	\begin{algorithmic}
		\STATE {\bfseries Input:} $\boldsymbol{x}$, $\Phi$, $\boldsymbol{y}$
		\STATE {\bfseries Output:} $\boldsymbol{x}^*$
		
		\FOR{$\boldsymbol{y}, \boldsymbol{x}$ in training data}
		
		\FOR{$\mathbf{W}_Q, \mathbf{W}$ in $\mathcal{F}_Q, \mathcal{F}$}
		\STATE $\mathbf{W}_Q \gets Q(\mathbf{W})$ 
		\ENDFOR
		
		\FOR{$\widetilde{\mathbf{W}}_Q, \widetilde{\mathbf{W}}$ in $\widetilde{\mathcal{F}}_Q, \widetilde{\mathcal{F}}$}
		\STATE $\widetilde{\mathbf{W}}_Q \gets Q(\widetilde{\mathbf{W}})$ 
		\ENDFOR
		
		\FOR{$k$ = 1 : $n_b$}
		
		\STATE $\boldsymbol{r}^{(k+1)} = \boldsymbol{x}^{(k)} - \rho {\mathbf{\Phi}^\textsc{T} (\mathbf{\Phi} \boldsymbol{x}^{(k)} - \boldsymbol{y})}$
		
		\STATE $\boldsymbol{x}^{(k)} = \widetilde{\mathcal{F}}_Q^{(k)}\Big(\sum_{i=1}^{p} \alpha_i \mathcal{P}_i(\mathcal{F}_Q^{(k)}(\boldsymbol{r}^{(k+1)}))\Big)$
		
		\ENDFOR
		
		\STATE Compute $\mathcal{L}_Q$ per Eq.~\eqref{Q-loss-function}
		
		\FOR{$\mathbf{W}$, $\widetilde{\mathbf{W}}$ in $\mathcal{F}$, $\widetilde{\mathcal{F}}$}
		\STATE $ \mathbf{W} \gets \mathbf{W} - \delta \nabla_{\mathbf{W}}\mathcal{L}_Q \, |_{\mathbf{W} = \mathbf{W}_{Q}}$
		\STATE $ \widetilde{\mathbf{W}} \gets \widetilde{\mathbf{W}} - \delta \nabla_{\mathbf{W}}\mathcal{L}_Q \, |_{\mathbf{W} = \widetilde{\mathbf{W}}_{Q}}$
		\ENDFOR
		
		\ENDFOR

	\end{algorithmic}
\end{algorithm}

\section{Quantized Deep-Unrolled Compressive Sensing Networks}
We now consider the effect of quantization of weights in the network. We consider $K$ bit quantization of the weights. We work with two versions of the filter weights -- the unquantized version and the quantized version. In the forward pass of Q-PAN, the full precision weights are mapped to the quantization levels, which are determined in a minimum mean-square error sense based on the weights obtained in the backpropagation update. The gradients of the loss function are computed with respect to the quantized weights. The weight updates happen on the full-precision version. The training strategy for Q-PAN models has been captured in Algorithm~\ref{alg:Q-PAN-Train}, where $Q(\cdot)$ denotes the quantization operator, defined in Fig. \ref{recon-block}. The quantization scheme is learnable. $Q(\boldsymbol{x}) = v\, \cdot \boldsymbol{b}$, where $\boldsymbol{b}$ has integer entries and $v$ is a scale factor. The scale factor is ``learned'' from the data $\boldsymbol{x}$. Effectively, we are learning to match the dynamic range of the quantizer to that of the data.\\

\begin{table*}[t]
	\centering
	\caption{PSNR [dB] comparison between quantized and unquantized networks on \textit{Set11} dataset between techniques from literature and Q-PAN$^+$ (3R) models.}
	\label{table:full-prec-results-on-set11-and-bsd68}
	\vskip 0.05in
	\begin{center}
		\begin{small}
			\begin{sc}
			\resizebox{0.8\linewidth}{!}{	\begin{tabular}{c||cccc||cccc}
			
					\toprule
					CS & & Recon& ISTA-& ISTA-& & 1-bit & 2-bit & 3-bit \\ 
					Ratio & SDA & Net & Net & Net$^+$ & PAN$^+$  & Q-PAN$^+$  & Q-PAN$^+$  & Q-PAN$^+$\\
					\midrule
					1$\%$  &  17.29 & 17.27 & 17.30 &  17.34 & 17.43 &  17.46 & 17.48 & {\bf17.49}\\
					
                    4$\%$  &  20.12 & 20.63 & 21.23 &  21.31	& {\bf21.83} &  20.95 & 21.20 & 21.47\\

					10$\%$ &  22.65 & 24.28 & 25.80 & 26.64	& {\bf 26.90} & 25.15 & 25.84 & 26.38\\
					
					\bottomrule 
				\end{tabular}}
			\end{sc}
		\end{small}
	\end{center}
	\vskip -0.1in
\end{table*}

\begin{table}[t]
	\centering
	\caption{Comparison between quantized and unquantized networks over average PSNR [dB] values on BSD68 dataset for different compression ratios. The PAN$^+$ models are of $3$R variant}
	\label{table:full-prec-results-on-set11-and-bsd68}
	\vskip 0.05in
	\begin{center}
		\begin{small}
			\begin{sc}
			\resizebox{\linewidth}{!}{	\begin{tabular}{c||c||ccc}
			
					\toprule
					CS& & &  1-bit  & 3-bit \\ 
					Ratio &  ISTA-Net$^+$ & PAN$^+$  & Q-PAN$^+$  & Q-PAN$^+$ \\
					\midrule
					10$\%$ &    25.33	& {\bf25.48} & 24.69 & 25.27\\
					4$\%$ &   22.17	& {\bf22.49} &  21.96 & 22.28 \\
					1$\%$ &    19.17	& {\bf19.21} & 19.15, & 19.21\\ 
					\bottomrule 
				\end{tabular}}
			\end{sc}
		\end{small}
	\end{center}
	\vskip -0.1in
\end{table}

The non-linear sparsifying transform $\mathcal{F}$ in Eq.~\eqref{eq: cs-composite} is modeled using a convolution network. In Q-PAN, a learnable quantization scheme maps the filter weights to lower-precision weights resulting in a quantized transform $\mathcal{F}_Q$. The sparsifying transform ($\mathcal{F}_Q$) is represented by conjunction of quantized convolution filters with ReLU. The update $\boldsymbol{x}$ at $(k+1)^\text{th}$ iteration with the quantized transform is given by
\begin{equation}
\label{quantized-update-eqn-composite}
\boldsymbol{x}_{k+1} = \arg \min_x J(x),
\end{equation}
where
\[ J(x)\ =\ \sum_{i=1}^{p} \frac{\alpha_i}{2\rho}\|\boldsymbol{x} - \boldsymbol{r}^{(k+1)}\|^2_2 + \sum_{i=1}^{p} \alpha_i g_i(\mathcal{F}_Q(\boldsymbol{x})),  \]
$\rho$ is the step-size and $\boldsymbol{r}^{(k+1)} = \boldsymbol{x}^{(k)} - \rho {\mathbf{\Phi}^T (\mathbf{\Phi} \boldsymbol{x}^{(k)} - \boldsymbol{y})}$. $\mathcal{F}_Q$ could be expressed in matrix form as $\mathcal{F}_Q(\boldsymbol{x}) = \mathbf{B}_Q \, \text{max}(\mathbf{A}_Q\boldsymbol{x}, \boldsymbol{0})$, where $\mathbf{A}_Q$ and $\mathbf{B}_Q$ correspond to the convolutional operators with quantized weights. Following the procedure outlined in Section 4, the structure of a layer in Q-PAN is modelled as follows:
\begin{equation}
\label{eq: QPAN-update-step}
    \boldsymbol{x}
    = \widetilde{\mathcal{F}}_Q
    \Bigg(
    \sum_{i=1}^{p} \alpha_i \mathcal{P}_i(\mathcal{F}_Q(\boldsymbol{r}))
    \Bigg),
\end{equation}
where the learnable parameters are $\{\lambda_i, a, \gamma, \mathcal{F}_Q,  \widetilde{\mathcal{F}_Q}\}$ subject to the conditions $\lambda_i > 0, \gamma > 1, a > 2$. The Q-PAN architecture is depicted visually in Figure~\ref{pipeline}.
Similarly, each layer of the Q-PAN$^+$ model is given by:
\begin{equation}
\label{eq:QPAN-plus-layer}
    \boldsymbol{x}
    = \boldsymbol{r} + 
    \mathcal{G}_{Q}(\widetilde{\mathcal{H}}_{Q}(\sum_{i=1}^{p} \alpha_i \mathcal{P}_i(\mathcal{H}_{Q}(\mathcal{D}_{Q}(\boldsymbol{r}))))).
\end{equation}
The weights of Q-PAN$^+$ are also quantized to $K$-bits. The loss function with weight quantization in the learning framework is given by 
\begin{eqnarray}
\mathcal{L}_Q &=&\frac{1}{N}{\sum^{n_b}_{i=1}\|\boldsymbol{x}^{n_l}_i - \boldsymbol{x}_i\|^2_2} \label{Q-loss-function} \\
&&+\gamma\, \frac{1}{N}\, {\sum^{n_b}_{i=1}\sum^{n_l}_{k=1}\|\widetilde{\mathcal{F}}_{Q}^{(k)}(\mathcal{F}_{Q}^{(k)}(\boldsymbol{x}_i)) - \boldsymbol{x}_i\|^2_2}. \nonumber 
\end{eqnarray}
Once the training is completed, only the quantized version of the weights is retained for inference. The architecture of one layer of Q-PAN is depicted in Fig.~\ref{recon-block}.

\section{Experimental Results and Discussion}
We validate the performance of our networks, PAN, PAN$^+$ and their quantized counterparts, namely, Q-PAN and Q-PAN$^+$ on natural image recovery and MR image reconstruction from compressed measurements. The results corresponding to PAN$^+$ and Q-PAN$^+$ are reported here. Due to space constraints, the results corresponding to PAN and Q-PAN are reported in the Supplementary Document. For performance assessment and comparison, we use peak signal-to-noise ratio (PSNR) and structural similarity index metric (SSIM) \cite{SSIM}.

\begin{figure}[htb]
\centering
	\includegraphics[width=0.85\linewidth]{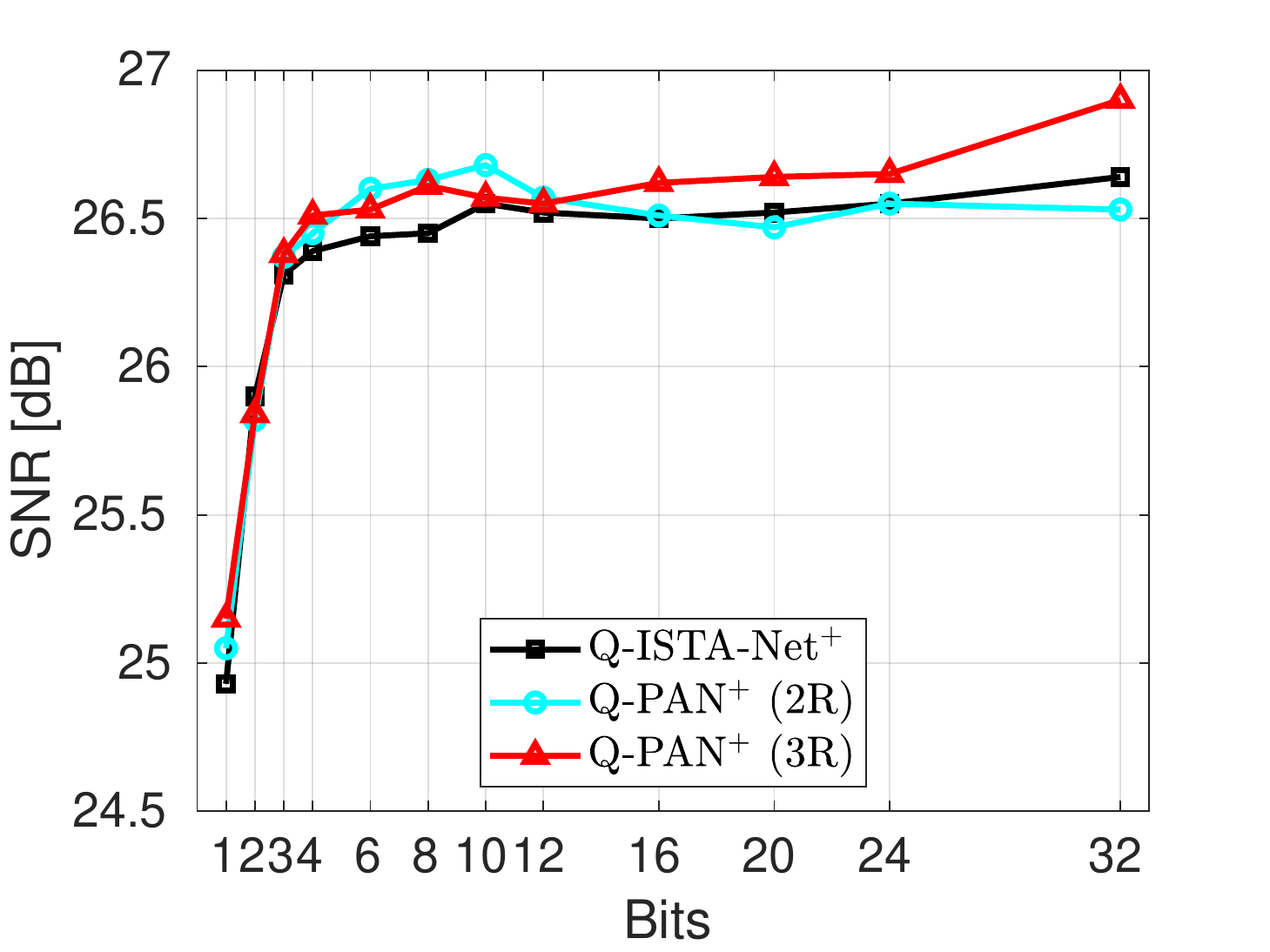}
	\caption{Comparative performance of Quantized ISTA-Net$^+$, Q-PAN$^+$ (2R) and Q-PAN$^+$ (3R) on Set11 dataset for CS Ratio = 10 with varying quantization bits}
	\label{PSNR-vs-bits-natural}
\end{figure}

\begin{figure*}[t]
\vskip 0.2in
\begin{center}
    \centerline{\includegraphics[width=1.0\linewidth]{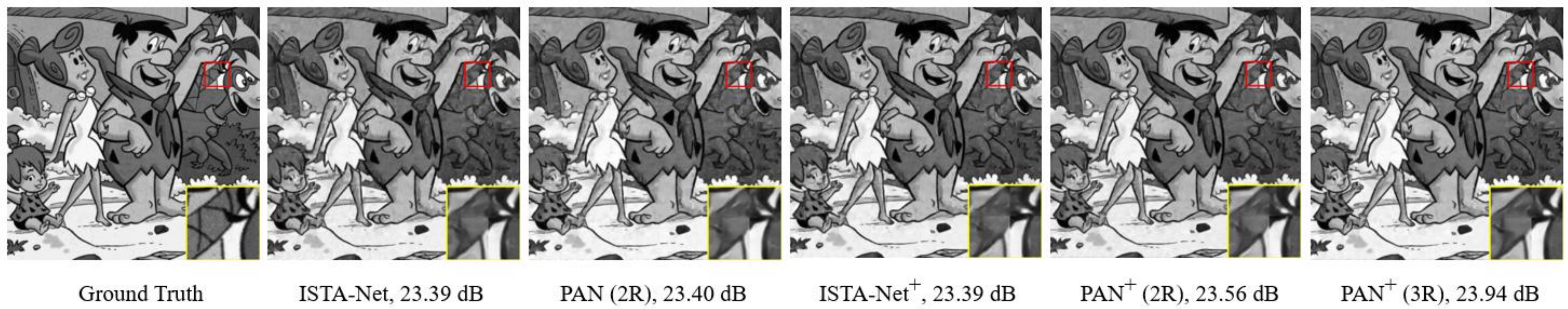}}
    \caption{Reconstruction results for {\it Flintstones} image based on 10\% compressed measurements.}
    \label{img:compare-flintstones}
\end{center}
\vskip -0.2in
\end{figure*}

\begin{figure}[htb]
\includegraphics[width = \linewidth]{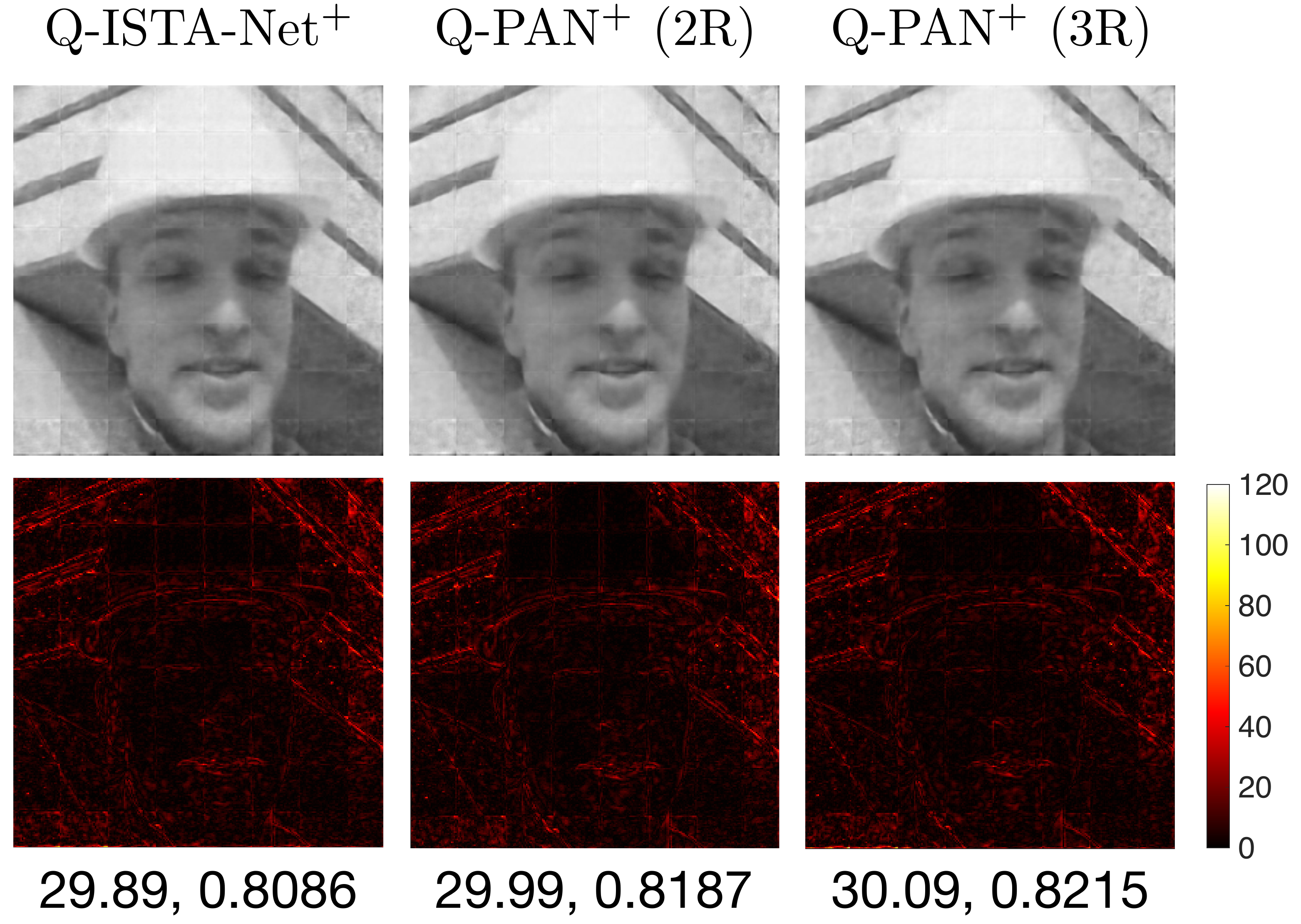}
\caption{Foreman (Set11) image reconstruction for CS ratio 10\% by the proposed 1-bit Quantized models. The bottom row visualizes the absolute difference between the ground truth and the reconstructed images. The numbers indicate the PSNR, SSIM values.}
\label{foreman-hot}
\end{figure}

\begin{figure}[htb]
    \centering
    \includegraphics[width = 0.85\linewidth]{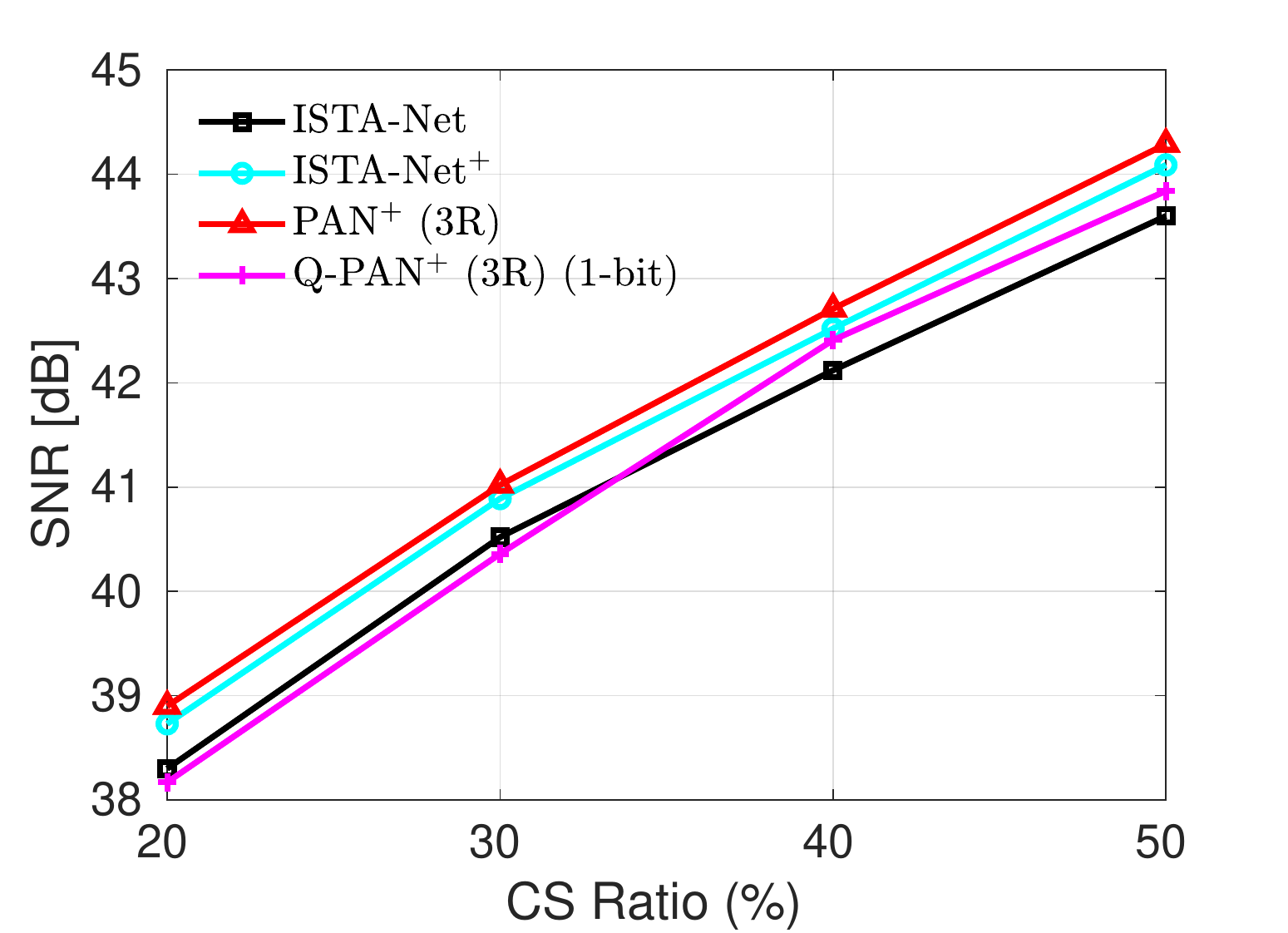}
    \caption{MR Image Reconstruction:~Performance comparison of Q-PAN$^+$ with the full-precision models, namely, PAN (Ours), ISTA-Net, and ISTA-Net$^+$ in terms of PSNR as a function of CS ratio. One can infer that Q-PAN$^+$ with $1$-bit quantization achieves image reconstruction performance on par with that of the full-precision models.}
    \label{fig:MRI_models}
\end{figure}

Several CS techniques \cite{TVAL3, D-AMP, IRCNN, SDA, ReconNet} are shown to be under-performing in comparison to ISTA-Net and ISTA-Net$^+$ \cite{ISTA-Net}. Therefore we compare the proposed PAN$^+$ and Q-PAN$^+$ models with ISTA-Net$^+$ and quantized ISTA-Net$^+$ (Q-ISTA-Net$^+$) in the context of compressed image reconstruction. ISTA-Net$^+$ uses single regularizer i.e. $\ell_1$ penalty, where as PAN$^+$ (3R) models employs two additional regularizers (SCAD and MCP). The performance improvement of PAN$^+$ (3R) over ISTA-Net$^+$ authenticates the superiority of three regularizers over one.

To ensure fair comparison, we use the same set of images as recommended in \cite{ISTA-Net} to train our models. The loss function, number of layers in the networks, the size and number of filters learned in the sparsifying function $\mathcal{F}$ are maintained the same for all the networks under consideration. The compressed measurements $\boldsymbol y$ are computed as $\boldsymbol y = \mathbf{\Phi} \boldsymbol{x}$, where $\mathbf{\Phi} \in \mathbb{R}^{m \times n} $ and $m$ varies depending on the compressive sensing (CS) ratio. The measurement matrix $\mathbf{\Phi}$ is obtained by orthonormalizing the rows of a random Gaussian matrix. In training our models, we set the batch size $n_b$ to $64$, and the number of layers $n_l$ as 9. In the transform $\mathcal{F}$, $n_f = 32$ filters of size $3 \times 3$ are learned in the first convolutional operation. The second convolutional operator learns $n_f = 32$ filters each of size $3 \times 3\times 32$.

The training data consists of 88,912 cropped image patches each of size 33 $\times$ 33. We trained the network models considering CS ratios of $1\%, 4\%$ and $10\%$ . The training and inference is carried out on a workstation with dual Intel® Xeon® Silver 4110 processors and RTX2080Ti GPU. The models are trained for 100 epochs and take approximately 9 hours. The PAN$^+$ (2R), Q-PAN$^+$ (2R) models employ two regularizers, namely the $\ell_1$ penalty and the MC penalty. The convex combination weights are fixed as $\alpha_1 = \alpha_2 = \frac{1}{2}$. The PAN$^+$ (3R), Q-PAN$^+$ (3R) models additionally use the SCAD penalty alongside $\ell_1$ and the MCP  with $\alpha_1 = \alpha_2 = \alpha_3 = \frac{1}{3}$.

The models are tested on the widely used Set11 \cite{ReconNet} and BSD68 \cite{BSD68} datasets, which contain 11 and 68 grayscale images respectively. We report the average PSNR over the test images in Table~\ref{table:full-prec-results-on-set11-and-bsd68}. 
To compare the effect of quantization on various models, we trained quantized versions of ISTA-Net$^+$ also. The variation of PSNR with the change in quantization bit-width is captured in Figure \ref{PSNR-vs-bits-natural}. We observe that Q-PAN$^+$ (3R) outperforms Q-ISTA-Net$^+$ consistently across different quantizations. The CS reconstructions by the proposed models are shown in Figs.~\ref{img:compare-flintstones} and \ref{foreman-hot} for visual inspection. From the zoomed-in figure (cf. Fig.~\ref{img:compare-flintstones}), one can infer that the PAN$^+$ (3R) model preserves the structure of the ground truth better, in comparison to the benchmark reconstructions. From the reconstruction error images (cf. Fig.~\ref{foreman-hot}) from the $1$-bit quantized models, one can observe that the Q-PAN$^+$ (3R) model contains fewer artefacts in its reconstruction.



\begin{table*}[ht]
	\centering
	\caption{Comparison of average PSNR [dB] on MR images between techniques from the literature and proposed models with three regularizers.}
	\label{table:full-prec-results-MRI-data}
	\vskip 0.05in
	\begin{center}
		\begin{small}
			\begin{sc}
					\resizebox{0.8\linewidth}{!}{\begin{tabular}{c||ccc||cccc}			\toprule
					CS & ADMM-& ISTA-& ISTA-& & 1-bit & 2-bit & 3-bit\\ 
					Ratio & Net & Net & Net$^+$ & PAN$^+$ & Q-PAN$^+$  & Q-PAN$^+$  & Q-PAN$^+$ \\
					\midrule
					
					20$\%$  & 	37.17 & 38.30 & 38.73, &	38.90 & 38.17 & $38.55$ & $\textbf{39.13}$ \\

					30$\%$  & 39.84 & 40.52 & 40.89 &	{41.02} & 40.36& $40.76$ & $\textbf{41.32}$ \\

					40$\%$   & 41.56 & 42.12 & 	42.52 &	{42.71} & 42.41 & $42.60$ & $\textbf{43.04}$ \\

					50$\%$   &  43.00 & 43.60 &	44.09 &	44.29 & 43.84 & $44.29$ & $\textbf{44.41}$\\				
					
					\bottomrule
				\end{tabular}}
			\end{sc}
		\end{small}
	\end{center}
	\vskip -0.1in
\end{table*}

\subsection{MR image reconstruction}
We evaluate the effectiveness of PAN$^+$ and Q-PAN$^+$ models on the CS MRI problem. The MR image reconstruction methods sample data in the Fourier space and adopt the CS theory to reconstruct images. During the reconstruction, the sensing matrix is obtained through a combination of under-sampling matrix ${\bf P}$ and a discrete Fourier Transform matrix ${\bf F}$ as in ${\bf \Phi = PF}$. We use the same training and testing brain medical images as in \cite{ISTA-Net}, and train models with $n_l = 11$ layers for CS ratios of $20\%, 30\%, 40\%, 50\%$. The comparison between ISTA-Net$^+$, PAN$^+$ (3R), and $1$-bit Q-PAN$^+$ (3R) techniques is summarized in Table~\ref{table:full-prec-results-MRI-data}. The experimental results shows that PAN$^+$ outperforms ISTA-Net$^+$ for all the compression ratios under consideration. Even when extremely quantized 1-bit representations are used, one can observe that the reconstruction performance degradation is at most 0.56 dB.

\section{Conclusion}
We proposed a multi-penalty formulation for the problem of compressed image reconstruction to learn data-driven analysis prior and developed proximal-averaged iterative shrinkage algorithm for solving it. We then unfolded the iterations to otain a neural network architecture. Making use of the knowledge that natural images are compressible in the residual domain, the enhanced network PAN$^+$ is proposed. We then incorporated a novel learnable quantization strategy into the unfolded networks and showed that the performance degradation even considering the extreme case of 1-bit quantization is less than 1 dB compared with the full precision case. The results present strong evidence that unfolded proximal-averaging networks with a quantizer incorporated into the loop in the learning stage offers competitive performance compared with the full-precision case. This makes a strong case for deployment of such networks on low-precision hardware.

\clearpage

\bibliographystyle{icml2021}
\bibliography{refs}

\end{document}